# Monocular 3D Object Detection for Proximity Prediction in Human-Machine Collision Warning Systems on Construction Sites

Yuexiong Ding[a,b], Xiaowei Luo[a,b*]

[a]Department of Architecture and Civil Engineering, City University of Hong Kong, Hong Kong, China

[b]Architecture and Civil Engineering Research Center, Shenzhen Research Institute of City University of Hong Kong, Shenzhen, China

**Abstract**

Monitoring workers' proximities to avoid human-machine collisions has aroused great concern in construction safety management. Existing methods are either too laborious and costly to apply extensively or lack spatial perception for accurate monitoring. This study proposes a novel framework for proximity monitoring using only an ordinary 2D camera to realize human-machine collision warning, which integrates a monocular 3D object detection model and a post-processing classification module to identify four proximity categories: ***Dangerous***, ***Potentially Dangerous***, ***Concerned***, and ***Safe***. A virtual dataset containing 22,000 images with 3D bounding box annotation is constructed and publicly released to facilitate the system development and evaluation. Experiments show that the implemented system is real-time and camera carrier-independent, achieving an F1 of roughly 0.8 within 50 meters. This study preliminarily reveals the potential and feasibility of proximity monitoring using only a 2D surveillance camera, providing a new promising and affordable way for early warning of human-machine collisions.

## 1. Introduction

Struck-by object is one of the leading causes of construction-related casualties [1]. Approximately 75% of struck-by fatalities involve heavy machines/vehicles/equipment [2]. According to the United States Department of Labor statistics, as many as 323 excavator-related accidents were recorded from 2011 to 2021, of which 179 cases caused at least one death [3]. Keeping a safe distance from heavy construction machines can reduce the occurrence of related accidents and casualties. For example, it is strictly specified that workers should never work within the swing radius of an operating excavator [4,5]. Real-time monitoring of workers' proximity for human-machine collision warning is an efficient way to avoid such accidents, which has been one of the critical issues to be addressed in construction safety management [6].

Various approaches have been proposed to monitor and estimate proximity to prevent collisions between workers and construction equipment. Human-based observation is commonly applied, which is costly and laborious. Many automated methods based on tag/wearable sensors thus have been utilized to improve management efficiency by monitoring the distance between machines and workers in real-time, such as the radio frequency identification (RFID) [1], ultra-wideband (UWB) [7], and Bluetooth [8,9]. However, these methods are also costly since they generally require a large number of sensors to tag each worker and machine. In addition, these tag sensors can not leave video or image records, which is inconvenient for subsequent management and accountability.

With artificial intelligence and deep learning development recently, computer vision-based (CV-based) methods for worker proximity alerts have gradually attracted significant attention due to automatic, non-invasive, and video recording advantages [6,10,11]. For example, Yan et al. [6] tried to estimate objects' 3D bounding boxes based on the predicted 2D boxes of objects' whole, top, side, and front (or back). However, based only on 2D annotated data, it is difficult for Yan's model to have spatial perception ability without the extra third-dimensional information for learning. Besides, estimating 3D bounding boxes based on the predicted 2D boxes might cause more significant errors due to the error accumulation from 2D results. Directly estimating objects' 3D positions and sizes using 3D object detection by learning from datasets with 3D annotations is a promising way to improve the situation. Unfortunately, there is no relevant dataset with 3D annotations in construction, resulting in few 3D object detection research and applications in this field.

To this end, the study aims to develop a lower-cost proximity monitoring system for real-time human-machine collision warning on construction sites, which only uses an ordinary 2D surveillance camera as the input source. Specifically, there are three objectives in the research: 1) construct and publicly release a virtual dataset with 3D bounding box annotations to alleviate the situation of lacking relevant data; 2) develop and train an end-to-end one-stage model for rapid 3D object detection from monocular 2D images; 3) define the proximity categories and implement a real-time monitoring system by integrating the trained 3D object detection model and a post-processing proximity classification module.

There are three novelties in this research. First, the constructed virtual image dataset with 3D annotations is automatically generated by coding automation programs on the Unity platform, which facilitates system development and verification and is expected to promote 3D CV development in construction. Besides, the proposed framework directly perceives workers' spatial status only from a single 2D image via monocular 3D object detection. Finally, the developed proximity monitoring system is based on a single ordinary 2D surveillance camera, with various superiorities of affordable, easy deployment and migration, and camera-carrier-independent.

The rest of the paper is organized as follows: Section 2 comprehensively reviews relevant works of proximity monitoring and 3D object detection in construction. Section 3 introduces the proposed framework for system development. Section 4 conducts the experiments and analyzes the performance of the implemented system, while Section 5 further discusses the characteristics and limitations of the implemented system. Finally, Section 6 concludes the work and its contributions to the area.

## 2. Literature review

### 2.1 Methods for proximity monitoring

*1) Sensor-based methods.* The key to proximity monitoring is to perceive the position information of objects in real-world space. In addition to human-based observation, the other methods can be categorized as sensor-based and CV-based. Tag sensors can determine their relative positions with others in space by sending and receiving signals in specific frequencies or bands. Many studies have used this feature for spatial positioning to identify proximity hazards on construction sites. For example, Teizer et al. [1] implemented a proximity security alarm system for construction workers and equipment operators using ultra-high frequency RFID sensors. Teizer and Cheng et al. [7] adopted UWB technology for real-time local positioning to monitor the proximity between workers and hazardous areas. Similarly, Park et al. [8] and Huang et al. [9] utilized Bluetooth technology for real-time positioning to achieve the same purpose. Sensor-based methods require each worker and equipment to be tagged with a sensor for mutual transmission and positioning, leading to an extra hardware cost to the contractor or manager. In addition, the union concerned may oppose tagging workers with sensors due to the potential health risks of close long-time electromagnetic radiation exposure [6]. Moreover, these tag sensors only leave digital signal records instead of videos or images, which is inconvenient for subsequent management and accountability.

*2) Hard vision-based methods.* In terms of spatial perception, CV-based methods can generally be divided into hard and soft methods. Hard methods construct the 3D scene by directly using depth cameras (also called RGB-D cameras) [12,13], which obtain depth information through integrated hardware based on structured light (e.g., Intel RealScene), time-of-flight (ToF, e.g., Microsoft Kinect), or camera array (e.g., Stereolabs ZED). However, these depth cameras are susceptible to outdoor light, limiting their degree of accuracy and measuring range outdoors [14]. In addition, high hardware and deployment cost hinder their further application on construction sites.

*3) Soft vision-based methods.* Compared with hard CV-based methods, deep learning-based methods are soft methods that estimate the relative position or distance between objects via the visual information captured by ordinary 2D RGB cameras. For example, Kim et al. [10] applied 2D object detection in the converted top view to identify proximity hazards. Yan et al. [6] tried to estimate objects' 3D bounding boxes based on the predicted 2D boxes of the objects' whole, top, side, and front (or back). Jeelani et al. [11] combined object detection, semantic segmentation, and simultaneous localization and mapping (SLAM) technologies to achieve worker positioning and proximity hazard detection. However, these studies were based on 2D data with only 2D annotations, making the trained models with little spatial perception ability. Besides, these methods estimated 3D information based on 2D predictions or ideal dimensions reduction. The cumulative errors caused by multi-step estimation or dimension reduction lead to low 3D prediction performance.

**2.2 3D object detection**

*1) Types of 3D object detection.* 3D object detection can achieve object positioning in the 3D space by providing objects' center point (x, y, z coordinates), size (length, width, height), and orientations. According to the type of data source, 3D object detection can be divided into multimodal data-based [15,16], point cloud data-based [17,18], binocular vision-based (or multi-view)[19,20], and monocular vision-based [21–23]. Methods based on multimodal data combine multi-source data from multiple RGB cameras, millimeter wave radar, LiDAR, etc., with state-of-the-art performance in the current automatic driving field. Followed by the point cloud data-based and binocular (or multi-view) vision-based methods, which are based only on point cloud data and images from two (or more) RGB cameras, respectively. Monocular vision-based methods rely only on a single image for 3D prediction. Though the performance is not the best, monocular 3D object detection still attracts great attention due to its rapid, low cost, and easy deployment. Compared with the additional deployment of expensive radar, LiDAR, or other RGB-D cameras, 3D space awareness based on a monocular 2D camera is a promising way to minimize the cost of achieving real-time proximity monitoring on construction sites.

*2) **Monocular 3D object detection and its situation in construction.*** Monocular 3D object detection is more complicated than the 2D case. The underlying problem is the inconsistency between the input 2D data and the output 3D predictions. Fortunately, it can be solved by geometric assumptions (such as flat-ground assumption), prior knowledge, or other information (such as depth estimation). Like 2D object detection, the monocular 3D object detection methods can also be divided into multi-stage and single-stage (or end-to-end) methods [24]. Multi-stage methods [25,26] usually transform 2D images to intermediate 3D representations (e.g., depth map, pseudo point clouds) and then make 3D predictions guided by the intermediate results. Though the multi-stage method has shown its prospects on many 3D datasets, its limitations are also evident. For example, Deep MANTA [25] relied on object-specific 3D shape assumptions. Pseudo-LiDAR [26] needed to deal with a large number of points generated in the reasoning process, leading to poor efficiency. The end-to-end models [21–23] eliminate these drawbacks by directly returning 3D localization information. Currently, relevant research and applications of 3D object detection almost focus on autonomous driving, but few in the construction field, mainly due to the lack of data with 3D bounding box annotation.

## 3. Methodology

Figure 1 shows the overall framework proposed in this study, which can be summarized into three phases: 1) data preparation, 2) modeling, and 3) system development. In the first stage, a virtual dataset with 3D bounding box annotations was automatically generated on the Unity platform. Then, the 3D object detection model pre-trained on the large public dataset was further trained and fine-tuned using the data generated in phase 1). Finally, the proximity monitoring system was implemented by integrating the 3D object detection model and a post-processing classification module, which identified predictions from the 3D detection model as predefined proximity categories.

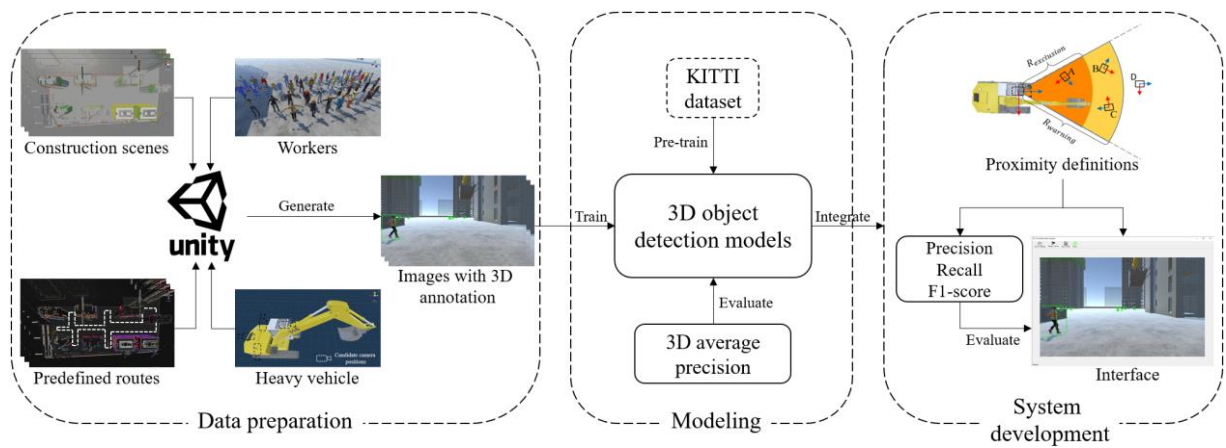

Figure 1. The proposed research framework.

### 3.1 Data construction

Though virtual/synthetic data can not completely replace real-world data, it helps achieve better performance based on a small amount of real-world data [27,28]. It is economical to use synthetic data to verify the effectiveness of the original idea or the prototype system at the initial stage, especially for research that is difficult or costly to obtain data. Therefore, this study constructed an image dataset captured from a monocular 2D surveillance camera attached to mobile construction machines (vehicles). Each image was annotated with workers' 3D bounding box annotations while captured. Based on the Unity platform and relevant packages [27], image capture and annotation were performed automatically through the developed Unity programs. For clarity, image capture and annotation processes are described separately below.

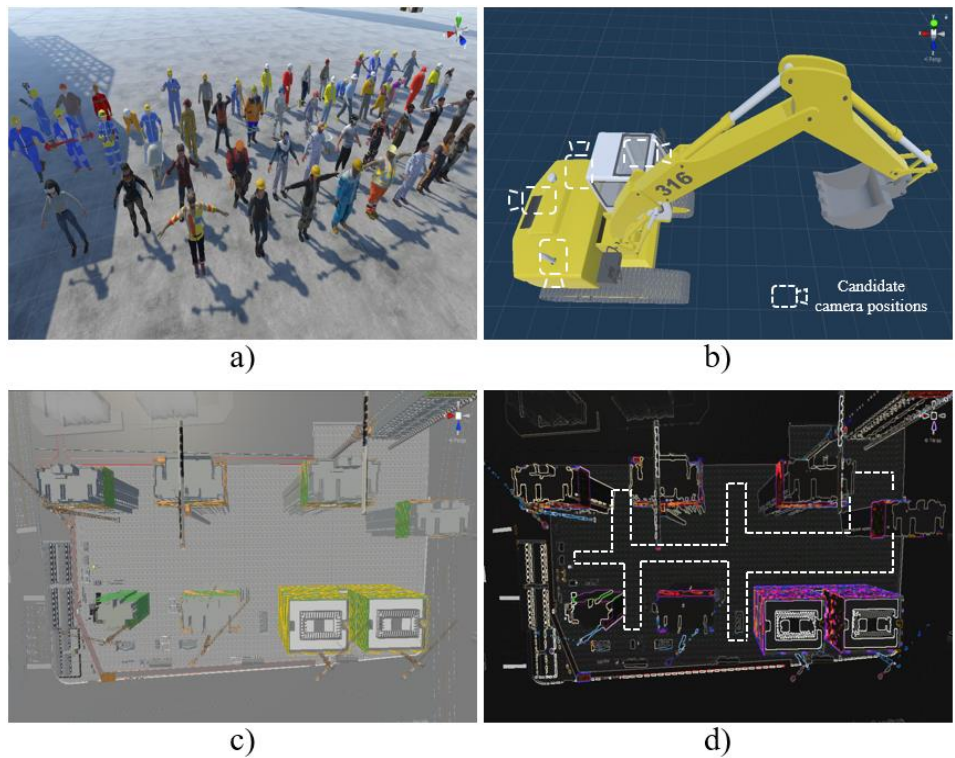

Figure 2. a) Prefabricated worker assets. b) Example of candidate camera positions for a vehicle. c) Example of the construction scene. d) Predefined driving route for the vehicle.

*1) Image capture.* Given a construction scene, the image capture process can be mainly divided into the following three steps:

- First is the determination of the worker instance set. Multiple worker instances are generated from the prefabricated worker asset library (Figure 2 a)) using the method of sampling with replacement. Each prefabricated worker asset in the library has the same probability of being selected, and the same asset can be selected multiple times to generate different worker instances.

- The second step is initialization and animation, which initializes the locations and orientations of worker instances and the heavy vehicle and makes them move in the scene at a predefined speed. The orientation of each worker/vehicle is adjusted at a random angle with a predefined probability to ensure random movement. In addition to random movement, the heavy vehicle can also move according to the predefined route (Figure 2 d)).

- The last is the periodical image capture. The attached cameras move synchronously with the vehicle and are programmed to save the visual information as an image at a given time interval (or frame rate). The camera position schemes on the construction machine might be multiple, and a simple but efficient scheme was considered in this study that attached cameras to the vehicle's front, rear, left, and right sides (Figure 2 b)).

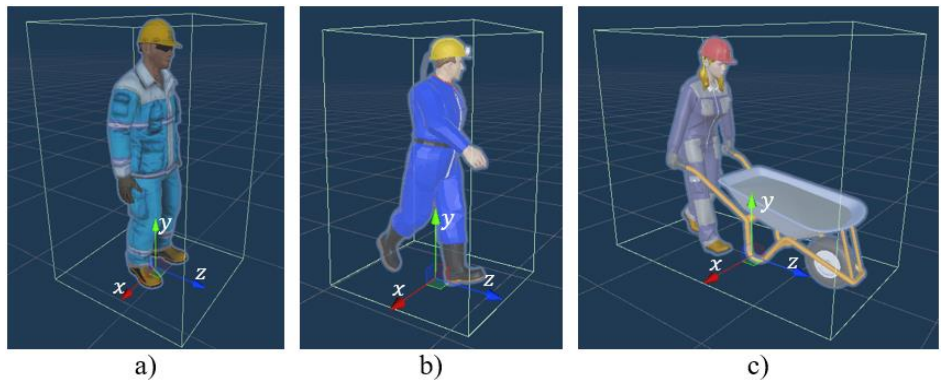

Figure 3. Examples of 3D bounding box annotation for the worker in different situations: a) ordinary worker, b) worker with inconsistent body and head orientation, and c) worker holding/carrying an object.

*2) 3D annotation.* Based on Unity's rich APIs, instances's spatial information (e.g., position, size, orientation) can be easily accessed and transformed. The 3D annotation in this study mainly focused on recording the camera status and 3D bounding information of workers in the current camera view. The recorded camera-related information included camera's intrinsic matrix and extrinsic parameters, such as translation and rotation, as shown in Table 1. The recorded 3D bounding box information mainly included the box's category, center location, size, and orientation, as shown in Table 2. The orientation of the 3D box in this study was defined as the orientation of the human torso (not the head), as shown in a) and b) in Figure 3.

The human torso's orientation was initialized to be always consistent with the z-axis under the global coordinate system in Unity. Note that the developed annotation programs regarded the worker and its holding/carrying objects as a whole, and the annotated 3D bounding box thus covered those objects, as shown in Figure 3 c).

Table 1. Main annotation fields related to the camera.

```
"sensor": {
   "sensor_id":        <str> -- Sensor ID.
   "translation":      <float, float, float> -- Sensor position(x, y, z) in meters, with respect to the global coordinate system.
   "rotation":         <float, float, float, float> -- Quaternion (w, x, y, z) of sensor orientation, with respect to the global coordinate system.
   "camera_intrinsic": <3x3 float matrix> -- Intrinsic camera calibration.
}
```

Table 2. Main annotation fields for one 3D bounding box.

```
[
    {
        "label_id":    <int> -- Integer ID of the label.
        "label_name":  <str> -- String label name (e.g., Worker).
        "instance_id": <int> -- Unique ID for each object instance.
        "translation": <float, float, float> -- Center location of the 3D bounding box in meters, with respect to the sensor's coordinate system.
        "size":        <float, float, float> -- Width, height, and length of the 3D bounding box in meters.
        "rotation":    <float, float, float, float> -- Orientation (w, x, y, z) of the 3D bounding box, with respect to the sensor's coordinate system.
    },
    {…}
]
```

## 3.2 Monocular 3D object detection

Considering the rapid response requirement of the early warning system, the full convolution one-stage (FCOS) model architecture [29] was considered in this study. The FCOS3D [22] and PGD [23], which developed on the FCOS architecture, have achieved advanced monocular 3D object detection performance on the KITTI [30] and Nuscens [31] datasets.

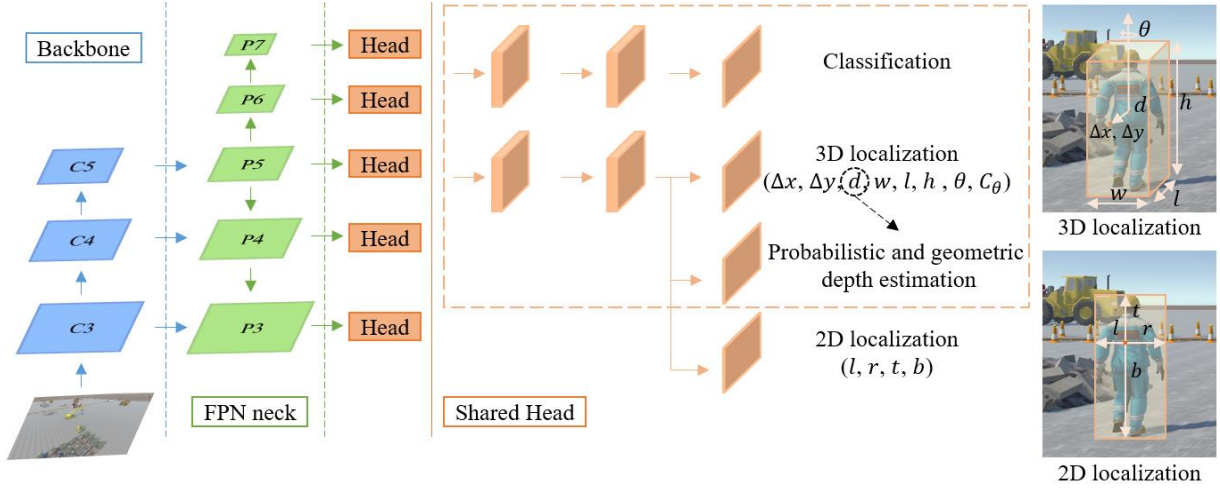

Figure 4. The general model architecture for monocular 3D object detection.

*1) Model architecture for monocular 3D object detection.* Figure 4 shows the general architecture of the FCOS3D and PGD models. The image feature is first extracted by the backbone network (e.g., Resnet101 [32]). Then multi-scale feature fusion is performed by the feature pyramid networks (FPN) [33]. Finally, output predictions via the shared head, including 3D bounding box attributes and auxiliary prediction results. In the 2D case, the model outputs 2D box attributes by regressing the distance of the point to the left, right, top, and bottom side, denoted as $l$, $r$, $t$, and $b$ in Figure 4. However, regressing the distance to the six faces of the 3D bounding box is redundant. On the contrary, the more direct way is to predict the commonly defined 7-DoF attributes, including the offset ($\Delta x$, $\Delta y$) from the box's center point to a specific foreground point, the depth ($d$) of the box's center point, the box's 3D dimensions ($w$, $l$, $h$), and its orientation. The prediction of the orientation is further divided into angle $\theta$ with period $\pi$ and 2-bin direction classification ($C_\theta$). The depth estimation of the FCOS3D was directly performed by convolution, while the PGD further introduced and combined the probabilistic and geometric depth estimation. The auxiliary predictions are generally used to calculate additional losses in training to provide more constraints for parameter learning. For example, the PGD model calculated the consistency loss from 3D to 2D using 3D and 2D localization predictions. Due to more accurate depth estimation, the PGD model outperforms the FCOS3D in many public 3D datasets. ,

*2) Evaluation for 3D object detection.* Like 2D cases, 3D object detection commonly uses the average precision (AP) metric for model evaluation, which extends the calculation of intersection over union (IoU) from the 2D bounding box to the 3D case. Specifically, this study adopted the 3D_AP11_easy_strict and 3D_AP11_easy_loose metrics defined by Geiger's research [30] to evaluate the performance of the 3D object detection model. "3D" means calculating IoU based on the 3D bounding box. "AP11" refers to the 11-point interpolated AP metric, first adopted in the VOC challenge for object detection evaluation [34]. "Easy" means only considering objects whose box height exceeds 42 pixels. "Strict" and "loose" represent different IoU thresholds to calculate the AP11, of which 0.5 for "strict" and 0.25 for "loose".

## 3.3 Proximity monitoring

As shown in Figure 5, the proximity of each worker to the machine/vehicle/equipment can approximately be turned into the distance between the worker and the attached camera, which can be calculated through its center coordinates ($x$, $y$, $z$) predicted by the 3D object detection model. In this case, the proximity is a continuous value measured by the distance between the worker and the camera. Practically, any proximity within a certain radius generally has the same degree of danger. Therefore, the proximity monitoring system in this study did not directly monitor the distance between the worker and the camera but further discretized the proximity by distance range to represent different danger levels. Based on the analysis above, the implementation principle of the proximity monitoring system becomes clear that it should at least contain a 3D object detection model and a post-processing module for proximity classification. In other words, the monitoring process mainly contains two loop steps: 3D object detection and proximity classification.

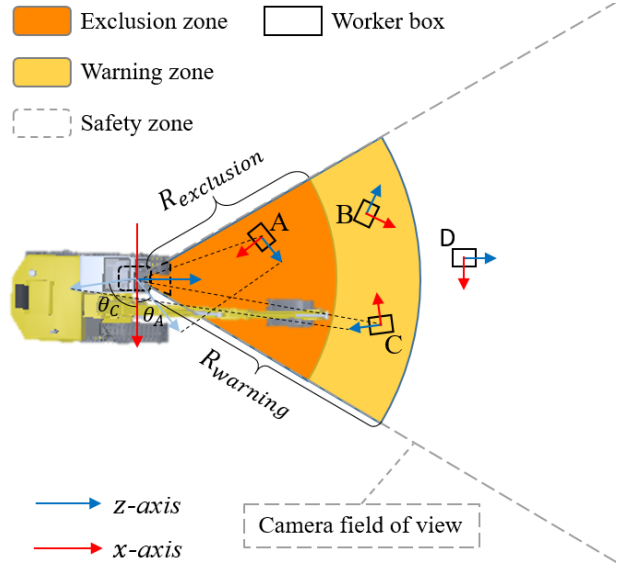

Figure 5. Demonstration of proximity identification within the camera field of view (FOV).

***Proximity category definition.*** Referring to and modifying from [9], there were four different proximity categories identified in this study: I) workers within the exclusion zone were labeled as ***Dangerous***, as worker A shown in Figure 5; II) workers within the warning zone with a tendency to approach the camera were labeled ***Potentially Dangerous***, as worker C shown in Figure 5; III) workers within the warning zone with a tendency away from the camera were labeled as ***Concerned***, as worker B shown in Figure 5; and IV) workers outside the warning zone were labeled as ***Safe***, as worker D shown in Figure 5. The worker's orientation in this study was defined as the angle between the worker's z-axis and the camera's x-axis. According to the rotation rule of the left-handed coordinate system, clockwise from the positive side of the x-axis to its negative side obtains a positive angle ($\theta_C$ in Figure 5), while counterclockwise gets a negative angle ($\theta_A$ in Figure 5). Based on the orientation definition, this study used the positive and negative attributes of the orientation angle to simply determine the approaching and leaving tendencies, respectively. Therefore, the four proximity categories can be formally represented by Equations (1), (2), (3), and (4), where $R_{exclusion}$ and $R_{warning}$ are customized radius of the exclusion and warning zones. $x_w$ and $z_w$ are the worker box's center coordinates in the *x* and *z* axes respective to the camera.

$$\text{I} = \left\{worker \Big| \sqrt{x_w{}^2 + z_w{}^2} < R_{exclusion}\right\} \quad (1)$$

$$\text{II} = \left\{worker \Big| \left(R_{exclusion} \leq \sqrt{x_w{}^2 + z_w{}^2} < R_{warning}\right) \text{ and } (\theta \geq 0)\right\} \quad (2)$$

$$\text{III} = \left\{worker \Big| \left(R_{exclusion} \leq \sqrt{x_w{}^2 + z_w{}^2} < R_{warning}\right) \text{ and } (\theta < 0)\right\} \quad (3)$$

$$\text{IV} = \left\{worker \Big| R_{warning} \leq \sqrt{x_w{}^2 + z_w{}^2}\right\} \quad (4)$$

*2) Proximity monitoring evaluation.* By turning into a typical classification task, the performance of the proximity monitoring system can be intuitively reflected by metrics of precision (5), recall (6), and F1 (7), where true positive (TP) is the number of positive samples correctly predicted by the model, false positive (FP) represents the number of positive samples incorrectly predicted, and false negative (FN) refers to the number of negative samples incorrectly predicted. Since each worker may have zero or more predicted boxes, a matching rule was introduced to match a unique prediction box for each ground-truth box. As shown in Equation (8), the $j^{th}$ prediction box is registered to the $i^{th}$ worker, where $j$ subjects to the max confidence $score_j$ among the predicted boxes whose IoU with the $i^{th}$ worker box ($IoU_j^i$) exceeding the customized threshold ($IoU^{thr}$). If all $IoU_j^i < IoU^{thr}$, then labels the $i^{th}$ worker as "None", indicating it does not be detected by the model. The uniquely matched boxes were then classified into corresponding proximity categories according to equations (1), (2), (3), and (4) to conduct the classification evaluation. Note that the $i^{th}$ worker labeled as "None" was identified as the "Unknown" proximity category.

$$Precision = \frac{TP}{TP + FP} \quad (5)$$

$$Recall = \frac{TP}{TP + FN} \quad (6)$$

$$F1 = 2 \times \frac{Precision \times Recall}{Precision + Recall} \tag{7}$$

$$Reg\_box^i = \begin{cases} Pred\_box^j, & \max_{j:IoU_j^i \geq IoU^{thr}}[..., score_j, ...] \\ None, & all\ IoU_j^i < IoU^{thr} \end{cases} \tag{8}$$

## 4. Experiments

### 4.1 Data overview

A total of 6 virtual construction scenes and 52 prefabricated worker assets were constructed in this study. These virtual scenes cover various construction activities that may involve large construction machines, such as earthwork excavation, concrete pouring, road laying, as shown in Figure 6. The worker asset library containes multiple poses, such as standing, walking, carrying, bending, and, squatting, as shown in Figure 2 a). To avoid overfitting, these prefabricated scene and worker assets were randomly divided into three exclusive subsets for training, validation, and test data generation, as shown in Table 3. During data generation, the excavator was selected as the mobile heavy machine to carry cameras moving around the scene. Then 2,000 consecutive images were captured and annotated from each scene at an interval of 0.2 seconds, as described in section 3.1. Referring to Table 3 for detailed information about the three sub-datasets and Figure 7 for examples of images with 3D bounding box annotation.

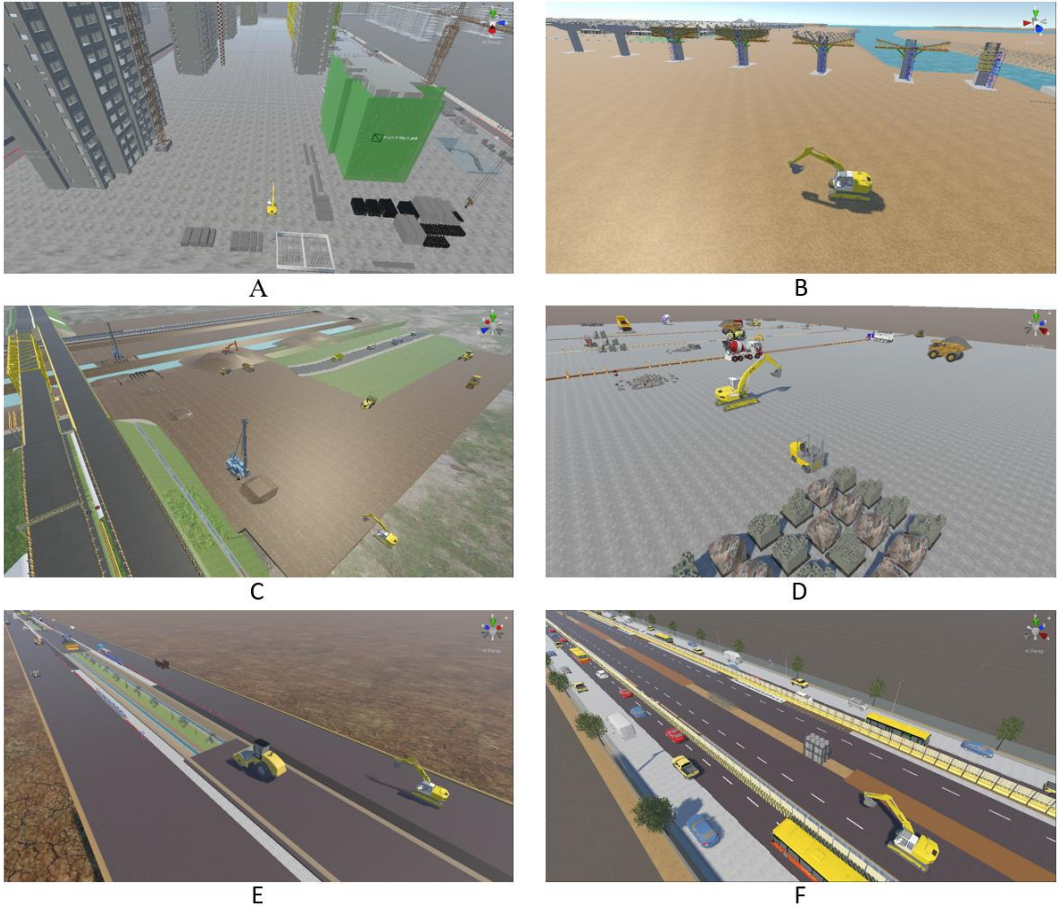

Figure 6. Overview of the considered construction scenes.

Table 3. Details of each sub-dataset.

| Dataset | Worker assets | Scene | Image | Cmara carrier | Move route | Camera |
|---|---|---|---|---|---|---|
| Training | 32 | B | 2,000 per scene | Excavator | Random | Front |
| | | A | | Excavator | Predefined | Front |
| | | E | | Excavator | Predefined | Front |
| Validation | 10 | C | 2,000 per scene | Excavator | Random | Front |
| | | F | | Excavator | Predefined | Front |
| Test_Front | 10 | D | 2,000 per camera | Excavator | Predefined | Front |
| Test_Rear | | | | Excavator | Predefined | Rear |
| Test_Left | | | | Excavator | Predefined | Left |
| Test_Right | | | | Excavator | Predefined | Right |
| Test_Front_Static | | | | Excavator | Static | Front |
| Test_Front_Truck | | | | Truck | Predefined | Front |

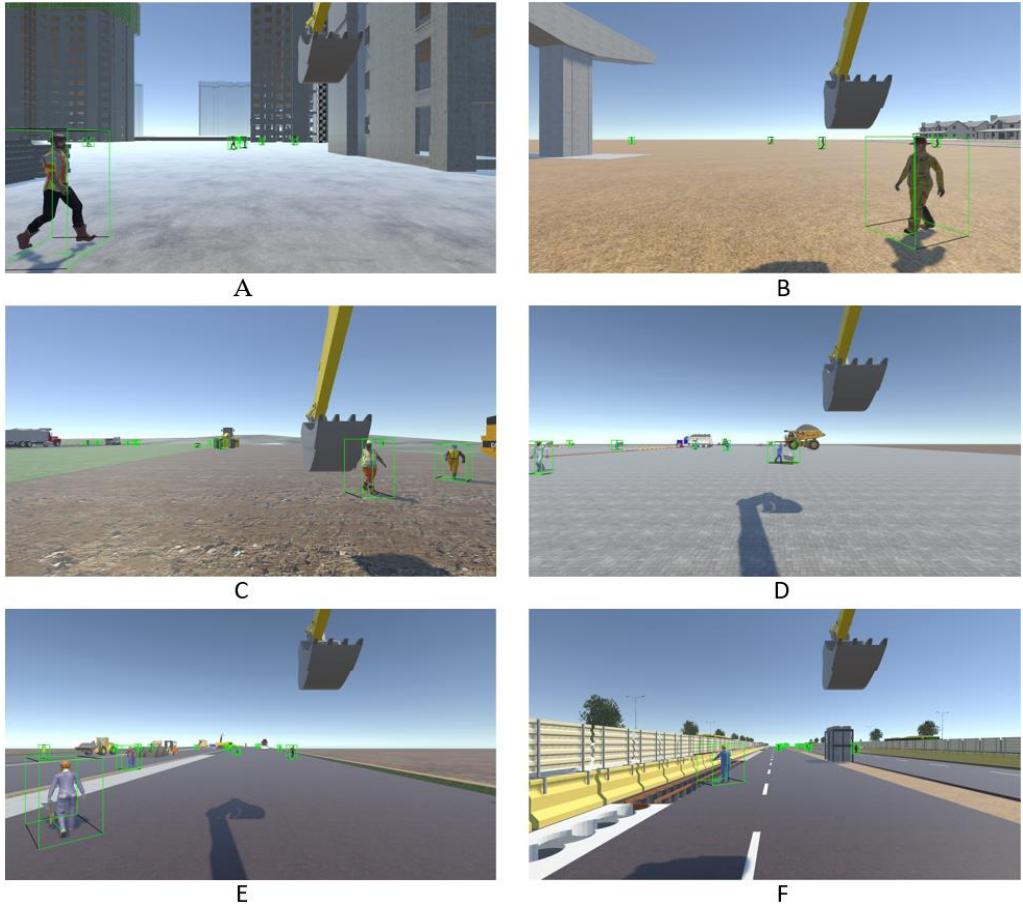

Figure 7. Sample images (from the front camera) with 3D annotations.

**4.2 Results of 3D object detection**

Three end-to-end structure-based models were considered for 3D object detection, including the SMOKE [21], FCOS3D [22], and PGD [23] models. These models were first trained on a large public 3D dataset KITTI [30] and then fine-tuned by the generated data using the stochastic gradient descent (SGD) strategy with a lower learning rate of 0.001. The input image size was set to 1920×1080, while the training batch size was only kept in one batch. All models were trained and tested in the environment of "Python 3.6 + RTX 2080 Ti 11G".

Table 4 compares the performance of the considered models on the Test_Front dataset. The PGD model outperforms the other two models when considering workers in all ranges, which achieves a loose AP of 49.35% and an mAP of 35.47%. The performances of the SMOKE and FCOS3D models are relatively similar because they focused only on developing a more efficient end-to-end architecture for 3D object detection and no specific designs for depth estimations. Based on FCOS3D's efficient architecture, however, the PGD put more effort into depth estimation since more accurate depth estimation makes a better 3D localization prediction [23], which is also significantly revealed by the test results in Table 4 that the PGD has 11.97% and 8.61% mAP increments compared with the SMOKE and FCOS3D.

Table 4. 3D object detection performance on the Test_Front dataset.

| Model | Range[a] (meter) | Instance[b] | 3D_AP11_easy_strict | 3D_AP11_easy_loose | **Mean** |
|---|---|---|---|---|---|
| SMOKE | (0, ∞) | 8,807 | 13.76 | 33.24 | 23.50 |
| FCOS3D | (0, ∞) | 8,807 | 15.72 | 38.00 | 26.86 |
| PGD | (0, ∞) | 8,807 | 21.58 | 49.35 | 35.47 |
| | (0, 10] | 230 | 38.23 | 74.35 | 56.29 |
| | (0, 20] | 868 | 37.51 | 74.94 | 56.23 |
| | (0, 30] | 1,747 | 30.45 | 64.88 | 47.66 |
| | (0, 50] | 3,761 | 22.75 | 51.58 | 37.16 |
| | (10, 20] | 638 | 36.30 | 73.16 | 54.73 |
| | (20, 50] | 2,893 | 13.16 | 42.57 | 27.87 |
| | (30, 50] | 2,014 | 11.43 | 35.74 | 23.59 |
| | (50, ∞) | 5,046 | 2.78 | 12.53 | 7.65 |

* a: Range (0,10] denotes filtering ground-truth and predicted boxes that meet the condition $0 < r \leq 10$, where $r$ is calculated by the $x$ and $z$ coordinates of the center point of the box: $r = \sqrt{x^2 + z^2}$.

* b: Instance denotes the number of ground-truth workers within the given range.

Another comparison experiment was conducted based on the PGD model by setting different distance ranges to explore the relationship between the detection performance and distance. The results in Table 4 show that: 1) the model performance decreases as distance increases; 2) The model has a stable and the best performance within a distance radius of 20 meters, with the highest loose AP of 75% and an mAP of 56%; 3) When distance exceeding 20 meters, the model performance begins to decline but still keeps a loose AP of more than 50% within 50 meters; 4) Beyond 50 meters, the model performance drops sharply with an mAP of only 7.56%, revealing that the model almost loses its detection ability for workers outside 50 meters away.

In short, though the PGD model performance decreases as distance increases, it performs well within 50 meters and keeps a stable and optimum status within 20 meters.

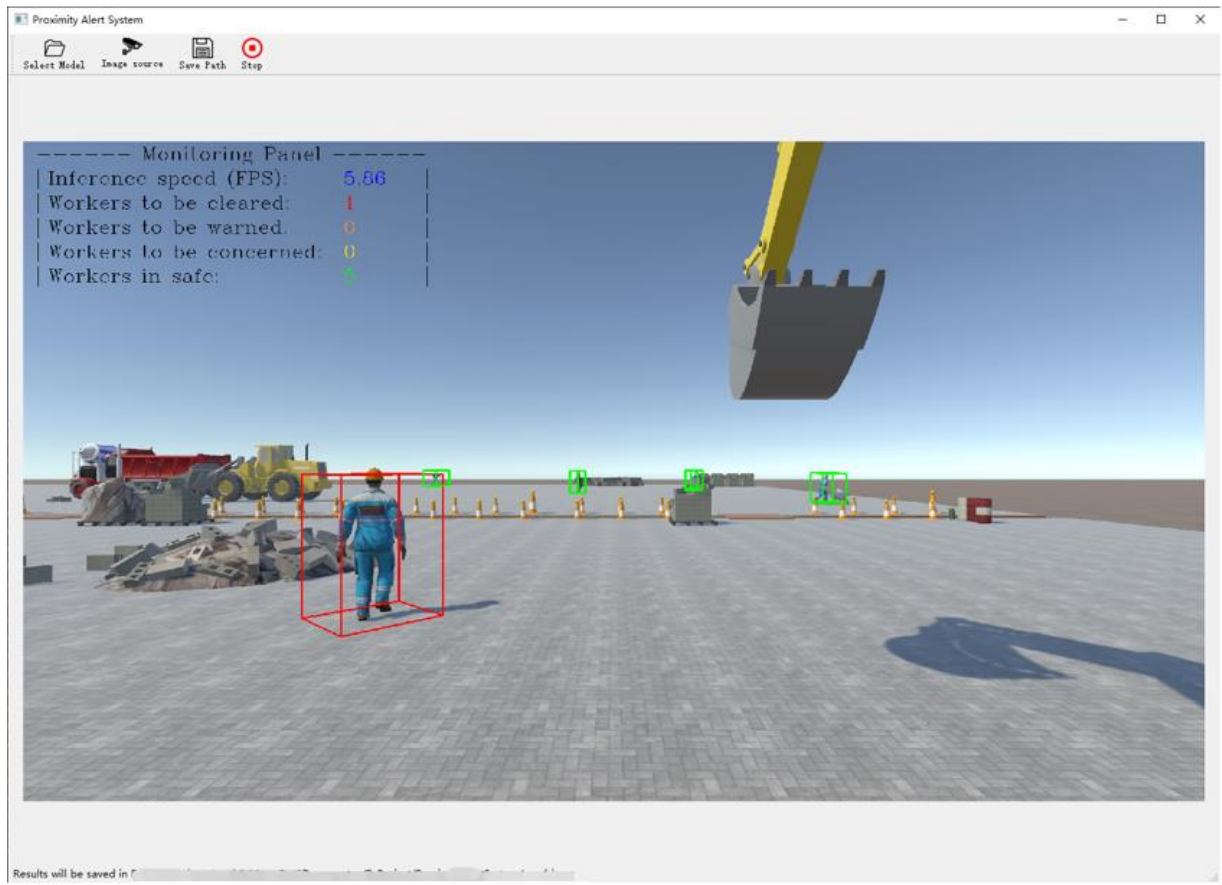

Figure 8. Interface of the developed proximity monitoring system.

## 4.3 System implementation and evaluation

According to the implementation principle described in section 3.3, a prototype system for worker proximity monitoring was realized through Python programming, whose inference core was developed by integrating the best 3D object detection model and a proximity classification module. Figure 8 shows the user interface of the system. For better visualization, different proximity categories were marked in different colors: red for workers in danger who need to be cleared immediately, orange for workers in potential danger who need to be warned, yellow for workers who need to be concerned, and green for workers in safety. A monitoring panel was designed and added to each processed frame to record the processing speed and the number of workers in each category. More monitoring examples are shown in Figure 9. In experiments on the Test_Front dataset, the developed system had an average processing speed of 5.8 frames per second (FPS) for the image size of 1920×1080, indicating the system can make early warnings within one second.

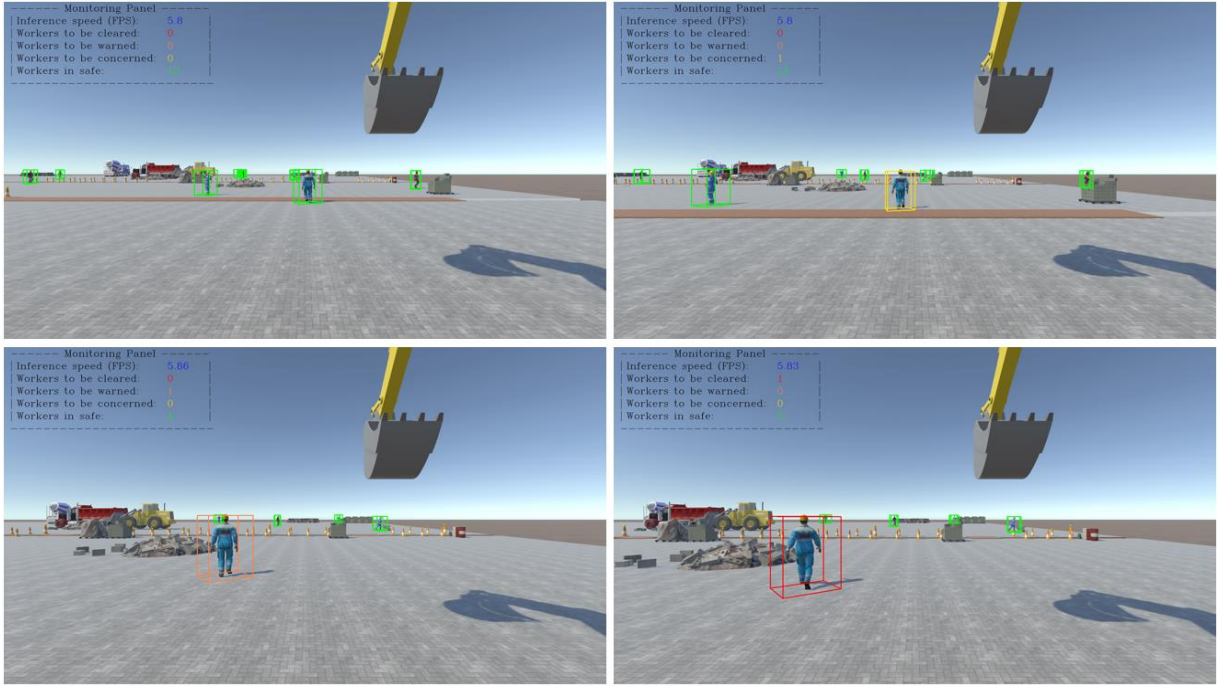

Figure 9. Some sample frames of the proximity detection.

The performance of the developed system was finally evaluated by the precision, recall, and F1 of the proximity classification. For demonstration, the customized parameters described in section 3.3 were tentatively set as follows: $R_{warning} = R_{exclusion}$, $IoU^{thr} = 0.25$. $R_{exclusion}$ is generally determined by the operation radius of the specific machine. For example, $R_{exclusion}$ can be set as the swing radius if the machine is an excavator [35]. However, there are too many different construction machines to set a unified $R_{exclusion}$. For general demonstration, this study chose 4, 7, and 10 meters to represent the operation radius of the small/compact, medium, and large machines after reviewing various types of excavators on sale [36]. The performance results are shown in Table 5. When without range limitation, the system achieves a satisfactory overall performance for different $R_{exclusion}$ settings, with mean F1 scores of 0.76, 0.75, and 0.72, respectively. Notably, the system achieves quite good scores in precision and recall for the most dangerous proximity (category I), with the highest F1 of 0.85, 0.9, and 0.94 for different $R_{exclusion}$.

Table 5. The performance of the developed system on the Test_Front dataset.

| Range | $R_{exclusion}$ | Proximity category | TP | TP+FP | TP+FN | Precision | Recall | F1 |
|---|---|---|---|---|---|---|---|---|
| (0, ∞) | 4 | I | 37 | 40 | 47 | 0.93 | 0.79 | 0.85 |
| | | II | 21 | 29 | 34 | 0.72 | 0.62 | 0.67 |
| | | III | 39 | 47 | 54 | 0.83 | 0.72 | 0.77 |
| | | IV | 4995 | 5000 | 8672 | 1.00 | 0.58 | 0.73 |
| | | **Mean** | / | / | / | 0.87 | 0.68 | 0.76 |
| | 7 | I | 80 | 81 | 96 | 0.99 | 0.83 | 0.9 |
| | | II | 71 | 101 | 131 | 0.7 | 0.54 | 0.61 |
| | | III | 169 | 227 | 209 | 0.74 | 0.81 | 0.78 |
| | | IV | 4702 | 4707 | 8371 | 1.0 | 0.56 | 0.72 |
| | | **Mean** | / | / | / | 0.86 | 0.69 | 0.75 |
| | 10 | I | 204 | 206 | 230 | 0.99 | 0.89 | 0.94 |
| | | II | 119 | 165 | 287 | 0.72 | 0.41 | 0.53 |
| | | III | 285 | 444 | 351 | 0.64 | 0.81 | 0.72 |
| | | IV | 4294 | 4301 | 7939 | 1.00 | 0.54 | 0.70 |
| | | **Mean** | / | / | / | 0.84 | 0.66 | 0.72 |
| (0, 50] | 4 | I | 37 | 40 | 47 | 0.93 | 0.79 | 0.85 |
| | | II | 21 | 29 | 34 | 0.72 | 0.62 | 0.67 |
| | | III | 39 | 47 | 54 | 0.83 | 0.72 | 0.77 |
| | | IV | 2,987 | 2,992 | 3,626 | 1.0 | 0.82 | 0.9 |
| | | **Mean** | / | / | / | 0.87 | 0.74 | 0.8 |
| | 7 | I | 80 | 81 | 96 | 0.99 | 0.83 | 0.9 |
| | | II | 71 | 101 | 131 | 0.7 | 0.54 | 0.61 |
| | | III | 169 | 227 | 209 | 0.74 | 0.81 | 0.78 |
| | | IV | 2,694 | 2,699 | 3,325 | 1.0 | 0.81 | 0.89 |
| | | **Mean** | / | / | / | 0.86 | 0.75 | 0.8 |
| | 10 | I | 204 | 206 | 230 | 0.99 | 0.89 | 0.94 |
| | | II | 119 | 165 | 287 | 0.72 | 0.41 | 0.53 |
| | | III | 285 | 444 | 351 | 0.64 | 0.81 | 0.72 |
| | | IV | 2,286 | 2,293 | 2,893 | 1.0 | 0.79 | 0.88 |
| | | **Mean** | / | / | / | 0.84 | 0.73 | 0.77 |

As revealed and inspired by Table 4, about 57% of workers are more than 50 meters away from the camera, and the 3D object detection model has the lowest performance in this range. Therefore, many workers 50 meters away were not accurately detected by the 3D object detection model, resulting in a low recall for proximity IV. In practice, monitoring the area within a radius of 50 meters is enough for almost all monitoring scenarios on construction sites. When the monitoring range narrows to 50 meters, as shown in Table 5, the recall for proximity IV is significantly improved, helping achieve higher mean F1 scores of 0.8, 0.8, and 0.77 for different $R_{exclusion}$. However, the recall for proximity II is the lowest regardless of the range size. Statistical analysis found that most ground-truth samples in proximity II were incorrectly classified as proximity III. According to equations (2) and (3), proximity II and III are the only two categories combining orientation angle for classification. Therefore, it can be inferred that the trained 3D object detection model was not good at orientation prediction, especially for angle direction classification, which tended to predict a negative direction.

The system was further evaluated using datasets captured by different machines, cameras, or moving schemes, as shown in Table 6. Though the 3D object detection model was trained on the data captured by the front camera attached to a mobile excavator, the integrated system achieves similar performances on datasets constructed by cameras on different machines (excavator and truck), sides (the front, rear, left, and right), and moving schemes (mobile and static). The performance fluctuations in Table 6 are within an acceptable and reasonable range, mainly caused by the different worker numbers within each radius range in different test datasets. Therefore, it is reasonable to conclude that the developed system can work efficiently using the cameras attached to any side of a mobile or static machine.

Table 6. System performance on different datasets captured by different cameras.

| Dataset | Mean precision | Mean recall | Mean F1 |
|---|---|---|---|
| Test_Front | 0.86 | 0.75 | 0.8 |
| Test_Rear | 0.83 | 0.68 | 0.74 |
| Test_Left | 0.78 | 0.69 | 0.73 |
| Test_Right | 0.83 | 0.72 | 0.76 |
| Test_Front _Static | 0.86 | 0.76 | 0.8 |
| Test_Front _Truck | 0.85 | 0.76 | 0.80 |
| **Mean** | 0.83 | 0.73 | 0.77 |

## 5. Discussion

Generally, early warning systems that can be applied in practice should be endowed at least with characteristics of high accuracy, generalization, and rapid response. According to the experimental results, the implemented proximity monitoring system has preliminarily possessed these features for field application:

- High accuracy. In terms of 3D object detection, the trained PGD model performed well within 50 meters, especially for those workers within 20 meters with a loose detection AP of up to 75%. High 3D detection performance helps obtain high proximity classification performance. Thus, within a radius of 50 meters, the system obtained an average F1 of roughly 0.8 for all categories. Notably, a higher F1 of about 0.9 was achieved for the most dangerous category (proximity I).

- High generalization. First, the system was developed with a high generalization for 3D object detection since the datasets used for model training, validation, and testing were exclusive, as described in section 3.1. Besides, the system was camera carrier-independent. As shown in Figure 7, though all images were captured by the cameras on the excavator, their content did not have much valuable visual information associated with the camera carrier for 3D object detection. The experimental results in Table 6 further prove the carrier-independent assertion.

- Rapid response. The developed system has achieved an inference speed of about 6 FPS for the image size of 1920×1080, indicating relevant warnings/measures theoretically could be taken as rapidly as 0.17 seconds later if any danger was detected. Response in one second has practically met the real-time requirement of many monitoring scenarios on construction sites. If only considering the inference speed, a smaller image size makes a faster speed. According to some additional experiments, the developed system achieved a speed of 16 FPS and 22 FPS for the image size of 960×540 and 480×270, respectively.

However, the developed system is still a prototype with several limitations. First, there is still much room for the monocular 3D object detection model to improve in the perspectives of inference accuracy and speed, especially the orientation prediction. Besides, the developed system can only respond/warn in time after dangerous proximity happens instead of in advance. In fact, relevant warnings can be raised much earlier if the probabilities of workers entering a specific zone are also calculated according to their orientation and movement velocities. Finally, the developed system is evaluated only on virtual synthetic data, and it would be better to evaluate the system further using real-world data. Therefore, constructing a real-world dataset with 3D annotations and developing a more advanced 3D object detection model with velocity predictions might be two promising directions to improve the performance of the proximity monitoring system.

## 6. Conclusions

This study proposes a novel and economical method to implement a real-time human-machine collision warning system by monitoring workers' proximities. To simplify proximity monitoring, the proposed method discretizes the straight distance between workers and machines into four different proximity categories: ***Dangerous***, ***Potentially Dangerous***,

*Concerned*, and *Safe*. By integrating a monocular 3D object detection model and a proximity classification module, a single 2D camera-based proximity monitoring system has been realized. The system was developed and evaluated on a virtual dataset containing 22,000 images with 3D bounding box annotations. The experimental results show that the best 3D object detection model has achieved a loose AP of 75% stably within the range of 20 meters and still kept a loose AP of more than 51% within 50 meters. Based on the well-trained 3D object detection model, the implemented proximity monitoring system has achieved an encouraging performance with an average F1 of 0.8 at a second-level response speed. The developed system has also been proven to be camera carrier-independent, which can be easily transferred and applied to any static or mobile machine equipped with a 2D camera. In future work, a new 3D object detection model with velocity predictions needs to be developed to help raise relevant warnings much earlier. In addition, a real-world dataset with 3D annotations should be created, and more experiments on real-world data should be conducted to achieve a more reliable system for practical application.

As a novel exploration, this study has several contributions to the area. First, the synthesized dataset significantly alleviates the shortage issue of datasets with 3D annotations in the construction area. In addition, the public release of the dataset might appeal to more 3D-related research and field applications in construction, which has been demonstrated in autonomous driving. Finally and most importantly, the successful implementation of the proximity monitoring system preliminarily proves the feasibility of perceiving proximity only using a single 2D camera, providing a low-cost approach to construction contractors/managers for automated human-machine collision warnings.

**Data availability statement**

The dataset constructed in this study can be publicly accessed on GitHub (https://github.com/dyxm/PM-HMCW).

**Acknowledgment**

The Shenzhen Science and Technology Innovation Committee Grant #JCYJ20180507181647320 and General Research Fund from Research Grant Council of Hong Kong SAR #11211622 jointly supported this work. The conclusions herein are those of the authors and do not necessarily reflect the views of the sponsoring agencies.